\documentclass[runningheads]{llncs}
\usepackage{graphicx}
\usepackage{amssymb}
\usepackage{longtable}
\usepackage{mathtools}
\usepackage{verbatim} 
\usepackage{dirtytalk}
\usepackage{tabularx}
\usepackage{mathtools}
\usepackage{comment}
\usepackage{amsmath}
\usepackage{multirow}
\usepackage{multicol}
\usepackage{float}
\usepackage{listings}
\usepackage{enumitem}
\usepackage{multirow}
\usepackage{xcolor}

\usepackage{rotating}

\newcolumntype{L}[1]{>{\raggedright\arraybackslash\textsc{}}p{#1}}
\floatstyle{ruled}
\newfloat{Listing}{thp}{lop}

\title{Discretizing Numerical Attributes: An Analysis of Human Perceptions}

\begin{document}

\author{Minakshi Kaushik\orcidID{0000-0002-6658-1712}\and Rahul Sharma\orcidID{0000-0002-9024-8768}\and Dirk Draheim\orcidID{0000-0003-3376-7489}}

\authorrunning{M. Kaushik}

\titlerunning{Discretizing Numerical Attributes:}

\institute{
Information Systems Group\\
Tallinn University of Technology\\
Akadeemia tee 15a, 12618 Tallinn, Estonia\\
\email{\{minakshi.kaushik,rahul.sharma,dirk.draheim\}@taltech.ee}}

\maketitle

\begin{abstract} 

Machine learning (ML) has employed various discretization methods to partition numerical attributes into intervals. However, an effective discretization technique remains elusive in many ML applications, such as association rule mining. Moreover, the existing discretization techniques do not reflect best the impact of the independent numerical factor on the dependent numerical target factor. This research aims to establish a benchmark approach for numerical attribute partitioning. We conduct an extensive analysis of human perceptions of partitioning a numerical attribute and compare these perceptions with the results obtained from our two proposed measures. We also examine the perceptions of experts in data science, statistics, and engineering by employing numerical data visualization techniques. The analysis of collected responses reveals that $68.7\%$ of human responses approximately closely align with the values generated by our proposed measures. Based on these findings, our proposed measures may be used as one of the methods for discretizing the numerical attributes.



\end{abstract}


\section{Introduction}
\label{sec1}

Various types of variables are available in real-world data. However, discrete values have explicit roles in statistics, machine learning (ML), and data mining. Presently, there is no benchmark approach to find the optimum partitions for discretizing complex real-world datasets. Generally, if a factor impacts another factor, in that case, humans can easily perceive the compartments or partitions because the human brain can easily perceive the differences between the factors and detect the partitions. However, it is not easy for a human or even an expert to find the appropriate compartments in complex real-world datasets.
In state-of-the-art, to find the optimum partitions of the numerical values, various discretization techniques have also been presented in the  literature~\cite{liu2002discretization,garcia2012survey,kotsiantis2006discretization}. However, the existing discretization techniques do not reflect best the impact of the independent numerical factor on the dependent numerical target factor. Moreover, no existing discretization approach uses numerical attributes as influencing and response factors. 

To find the cut-points for the cases of two-partitioning and three-partitioning, we have proposed two measures~\emph{Least Squared Ordinate-Directed Impact Measure} (LSQM) and~\emph{Least Absolute-Difference Ordinate-Directed Impact Measure} (LADM)~\cite{impact2021}. These measures provide a simple way to find partitions of numerical attributes that reflect best the impact of one independent numerical attribute on a dependent numerical attribute. In these measures, we use numerical attributes as influencing and response factors to distinguish them from the existing approaches.

In this paper, the outcome of ~\emph{LSQM} and ~\emph{LADM} measures are compared with the human-perceived cut-points to assess the accuracy of the measures. We use numerical attributes as influencing and response factors to distinguish them from the existing approaches. A series of graphs with different data points are used to collect the human responses. Here, data scientists, ML experts and other non-expert persons are referred to as humans.

The idea of this research emerged from the research on partial  conditionalization~\cite{draheim2017generalized,Dra19}, association rule mining (ARM)~\cite{sharma2020expected,bigdata} and numerical association rule mining (NARM)~\cite{srikant1996,kaushik2020potential,kaushik2021systematic}. These papers discuss the discretization process as an essential step for NARM. Moreover, research on discretizing the numerical attributes is an essential step in frequent itemset mining, especially for quantitative association rule mining (QARM)~\cite{srikant1996}. 

In the same sequence, we have also presented a tool named Grand report~\cite{peious2020grand} and a framework~\cite{rahul} for unifying ARM, statistical reasoning, and online analytical processing. These paper strengthens the generalization of ARM by finding the partitions of numerical attributes that reflect best the impact of one independent numerical attribute on a dependent numerical attribute. Our vision is to develop an ecosystem to generalize the ML approaches by significantly improving the ARM from different dimensions.

The paper is organized as follows. In Sect.~\ref{sec2}, we delve into the discussion of related work concerning discretization and its connection with human perception. This section aims to provide a comprehensive overview of prior research and studies that have explored the topic from different angles. In Sect.~\ref{sec3}, we explain the motivation for conducting this study. Sect.~\ref{sec4} describes the ~\emph{LSQM} and ~\emph{LADM} measures. Then, we describe the design of the experiment in Sect.~\ref{sec5}. In  
 Sect.~\ref{sec6}, we present the analysis and results.  The conclusion and future work are given in Sect.~\ref{sec7}.

\section{Related Work}
\label{sec2}
Based on human perception evaluation and different discretization techniques, we discuss the related work in the direction of discretization and clustering techniques and human perception.

\subsection{Discretization}
Many data mining and ML algorithms are not designed to work with numeric attributes and instead require nominal attributes as input data~\cite{garcia2012survey}. In order to convert numeric attributes into nominal attributes, different discretization methods have been employed as a pre-processing measure. Discretization techniques divide the range of a numeric attribute into $n$ intervals, which are determined by $n-1$ cut-points. A variety of discretization methods are available in the literature~\cite{garcia2012survey},\cite{kotsiantis2006discretization} and \cite{liu2002discretization}. Dougherty et al.~\cite{dougherty1995supervised} compared and analyzed discretization strategies along three dimensions: global versus local, supervised versus unsupervised, and static versus dynamic. Static approaches discretize each attribute separately, whereas dynamic methods conduct a search through space to find interdependencies between features. Liu et al.~\cite{liu2002discretization} performed a systematic study of existing discretization methods and proposed a hierarchical framework for discretization methods from the perspective of splitting and merging. The unsupervised static discretization method, such as equal-width, uses the minimum and maximum values of the continuous attribute and then divides the range into equal-width intervals called bins. In contrast, the equal-frequency algorithm determines an equal number of continuous values and places them in each bin~\cite{catlett1991changing}.

The RUDE (Relative Unsupervised Discretization) algorithm~\cite{lud2000relative} performs the discretization of numerical attributes in three steps: pre-discretizing, structure projection, and merging split points. In 2010, Joita~\cite{joicta2010unsupervised} proposed an unsupervised method for selecting the initial cluster centers for the clustering of a one-dimensional vector of real-valued data. This method was based on a k-means clustering algorithm and can be used in single-attribute discretization. In 2013, Dietrich et al.~\cite{dietrich2013smoothed} also proposed a method for obtaining cut-points that is more intuitive for human users. This smoothed discretization approach works as a post-processing step to obtain the intervals after using an arbitrary traditional discretization approach. The authors also proposed two measures,~\emph{distance-based deviation measure} and~\emph{instance-based deviation measure}, for comparing the original discretization method cut-points with modified cut-points. The discretization cut-points were computed for each training dataset for the three general discretization methods: equal-frequency~\cite{catlett1991changing} discretization, entropy-based discretization~\cite{fayyad1993multi},~\cite{quinlan1986induction} and Chi2~\cite{Chi2} discretization. After that, the introduced smoothing approaches were used with distance-based and instance-based modification measures to get the desired results. There is another similar work, called the best piecewise constant approximation~\cite{konno1988best}, which deals with approximating a single variable function. Still, it is different because we are not using signals, and our primary focus is on data sets that use several data points for one value of the influencing factor. Eubank used the population quantile function as a tool to show the best piecewise constant approximation problem~\cite{eubank1988optimal}. Later Bergerhoff~\cite{bergerhoff2019algorithms} suggested a method for finding optimal piecewise constant approximations of one-dimensional signals using particle swarm optimization.

\subsection{Human perceptual evaluation}
Discretization approaches are usually evaluated based on their mathematical backgrounds. However, we are the first to assess the discretization measures by considering human perceptions. 

In the state of the art, many studies have used human perception to evaluate various techniques. However, they are not completely related to discretization. 
Tatu et al.~\cite{Tatu2010} proposed a preliminary investigation of human perception using visual quality criteria for multidimensional data. The authors conducted a user study to examine the relationship between human cluster interpretation and the measurements automatically retrieved from 2D scatter plots. Etemadpour et al.~\cite{Etemadpour2014} conducted a perception-based evaluation of high-dimensional data where humans were asked to identify clusters and analyze distances inside and across clusters.  Demiralp et al.~\cite{demiralp2014learning} used human judgments to estimate perceptual kernels for visual encoding variables such as shape, size, color, and combinations. The experiment used Amazon's Mechanical Turk platform, with twenty Turkers completing thirty MTurk jobs. In~\cite{abbas2019}, a new visual quality measure (VQM) based on perceptual data was proposed to rank monochrome scatter plots. This experiment collected perceptual data from human subjects, and the best clustering model was chosen to create a perceptual-based VQM of grouping patterns. Similarly, a study by Aupetit~\cite{Aupetit2019} analyzed and compared clustering algorithms through the lens of human perception in 2D scatter plots. The primary focus of the authors was to evaluate how accurately clustering algorithms aligned with the way humans perceive clusters. The authors evaluated Gaussian Mixture Models, CLIQUE, DBSCAN, Agglomerative Clustering methods, and 1437 variations of k-means on the benchmark data. Our work is also related to considering human perceptions for evaluating our proposed~\emph{LSQM} and~\emph{LADM} measures for discretizing numerical attributes.

\section{Motivation}
\label{sec3}
Real-world data sets contain real or numerical values frequently. However, many data mining and ML approaches need discrete values. For years, obtaining discrete values from numerical values has been a complex and ongoing task. The main issue with the discretization process is obtaining the perfect intervals with specific ranges and numbers of intervals. In state of the art, several discretization approaches such as equi-depth, equi-width~\cite{catlett1991changing}, MDLP~\cite{fayyad1993multi}, Chi2~\cite{Chi2}, D2~\cite{catlett1991changing}, etc. have been proposed. However, determining the most effective discretizer for each situation is still a challenging problem. The existing methods for discretizing numerical attributes are not automated and require expert knowledge; therefore, there is a need to develop an automated and formal measure for finding the optimal partition of numerical attributes.

In~\cite{impact2021}, we presented an order-preserving partitioning method to find the partitions of numerical attributes that reflect best the impact of one independent numerical attribute on a dependent numerical attribute. In extreme cases (such as step-functions), humans can easily visualize the perfect partitions and even the number of compartments. However, in distinct cases, the ideal partition range depends on the perception of data experts. In state of the art, no investigation is available to understand the human perception of partitioning. Moreover, the current literature provides a comparison of discretization methods and compares their results. In this paper, we take a different approach to compare the human perception of discretization with the outcome of the proposed discretization method. We aim to visualize the differences between the outcomes of the proposed methods and the human perception of discretization.

\section{Impact Driven discretization Method}
\label{sec4}
In the ~\emph{Impact driven discretization method}~\cite{impact2021}, we perform discretization on the independent numerical attribute using order-preserving partitioning to understand its impact on the numerical target attribute. The method involves creating a total of $(k-1)$ cut-points, with $k$ being the number of partitions recommended by the user.
Below are two measures introduced in the paper~\cite{impact2021}.

\subsection{The LSQM Measure}
\label{LSQM}
The~\emph{LSQM} measure operates by initially computing the squared difference between the
$y$-value of each data point and the average of 
$y$-values within the current partition. This measure maintains the order of the independent variable by considering the values of data points, ensuring that the values within one partition are consistently lower than those in the subsequent partition.
After summing up the squared differences of the several partitions,~\emph{LSQM} retrieves the minimum values as cut-points.

\begin{definition}[Least Squared Ordinate-Directed Impact Measure]\label{LSODdef}
\normalfont
\hfill \\ Given $n \! \ge \! 2$
real-valued data points 
$(<x_i, y_i>)_{1 \leq i \leq n}$, we define the \emph{least squared ordinate-directed impact measure} for 
$k$-partitions (with 
$k \! - \! 1$ \emph{cut-points}) as follows: 

\begin{equation}
\label{eq1}
\mathop{min}_{i_0=0< i'_1<...<i'_{k-1}<i'_k=n}
 \sum_{
\scriptsize
\begin{array}{c}
j=1\\[-0.5mm]
\end{array}
}^{k} \hspace{0.5mm}
\sum_{i'_{j-1} < i" \leq i'_j}
(y_{i"}-\mu_{i'_{j-1} < \phi \leq i'_j})^2
\end{equation}

where the \emph{average of data values in a partition} 
$\mu_{a < \phi \leq b}$ between indexes
$a$ and
$b$ ($a < b \le n$)  is defined as
\begin{equation}
    \mu_{a < \phi \leq b} =\frac{\sum\limits_{a < \phi \leq b} y_\phi} {b-a}
\end{equation}

\end{definition}

In (\ref{eq1}), we have that 
$i'_j$ is the highest element in the $j$-\emph{th} partition, where~\emph{highest element} means the data point with the highest index.

Indeed, the ~\emph{LSQM} (Least Squares Ordinate-Directed Impact Measure) measure may appear similar to the $k$-means clustering algorithm on the surface, as both involve partitioning data into clusters. However, they differ significantly in their underlying principles and applications.


$k$-means clustering is primarily an unsupervised ML technique employed for the task of clustering data points into groups or clusters, with each data point assigned to the cluster whose centroid is closest to it in terms of a chosen distance metric, often the Euclidean distance. Euclidean distance metric calculates dissimilarity between data points, which involves measuring the geometric distance between vectors $X$ and $Y$. The primary goal of $k$-means is to minimize the sum of squared distances between data points and their assigned cluster centroids, and it finds applications in various domains, including customer segmentation, image compression, and data reduction. 
However, $k$-means' effectiveness is influenced by the initial random selection of cluster centers, which can lead to different clustering results depending on the initialization.

 
In contrast,~\emph{LSQM} is a specialized method designed specifically for discretizing numerical attributes. Its core objective is to partition a numerical attribute into intervals while preserving the order of data points within those intervals. ~\emph{LSQM} achieves this by measuring the squared difference between the values of data points and the average of values within each partition, aiming to minimize the sum of squared differences.
Unlike $k$-means,~\emph{LSQM} is not highly dependent on the initial point chosen to start the partitioning process, making it robust in this regard. ~\emph{LSQM} is primarily employed in data preprocessing tasks related to data mining, enhancing the quality of numerical attribute discretization.

In summary, $k$-means clustering is a versatile and widely used clustering algorithm with applications across various domains, focusing on minimizing the squared distances between data points and cluster centroids. On the other hand,~\emph{LSQM} serves a specific purpose in discretizing numerical attributes while maintaining the order of data points, making it particularly valuable in data preprocessing for data mining tasks.

\subsection{The LADM Measure}
\label{LADM}
For the~\emph{LADM} measure, we take the sum of the absolute differences of the several partitions.

\begin{definition}[Least Absolute-Difference Ordinate-Directed Impact Measure]\label{LAODdef}
\normalfont
\hfill \\ Given $n \! \ge \! 2$
real-valued data points 
$(<x_i, y_i>)_{1 \leq i \leq n}$, we define the \emph{least absolute-difference ordinate-directed impact measure} for 
$k$-partitions (with 
$k \! - \! 1$ \emph{cut-points}) as follows: 

\begin{equation}
\label{eq2}
\mathop{min}_{i_0=0< i'_1<...<i'_{k-1}<i'_k=n}
 \sum_{
\scriptsize
\begin{array}{c}
j=1\\[-0.5mm]
\end{array}
}^{k} \hspace{0.5mm}
\sum_{i'_{j-1} < i" \leq i'_j}
|y_{i"}-\mu_{i'_{j-1} < \phi \leq i'_j}|
\end{equation}
where the \emph{average of data values in a partition} 
$\mu_{a < \phi \leq b}$ between indexes
$a$ and
$b$ ($a < b \le n$)  is defined as
\begin{equation}
    \mu_{a < \phi \leq b} =\frac{\sum\limits_{a < \phi \leq b} y_\phi} {b-a}
\end{equation}

\end{definition}

\section{Experimental Design}
\label{sec5}

\begin{figure}
\centering
\includegraphics[width=\textwidth]{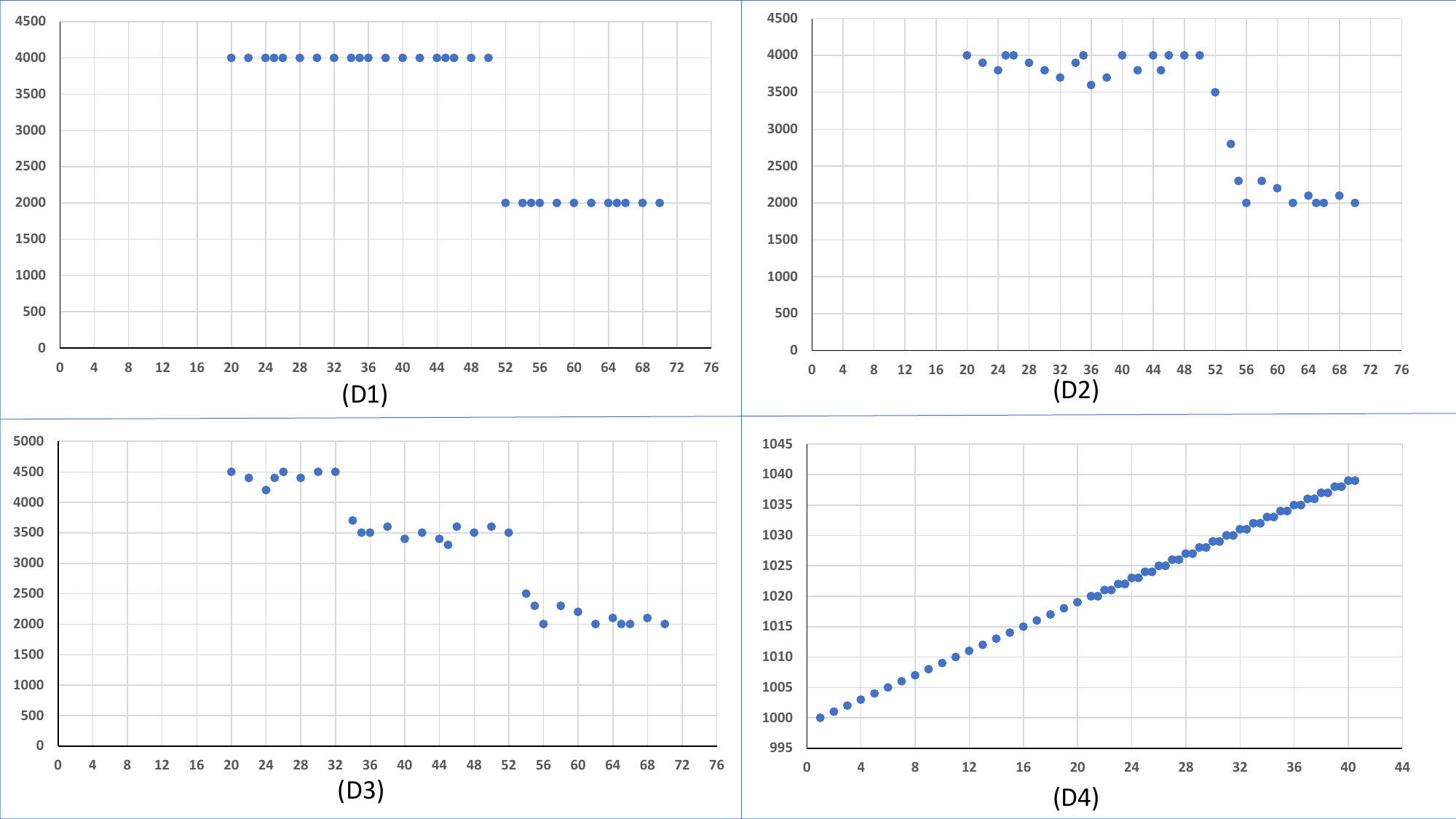}
\caption{Graphs for datasets D1 to D4.}
\label{fig1}
\end{figure}

\begin{figure}
\centering
\includegraphics[width=\textwidth]{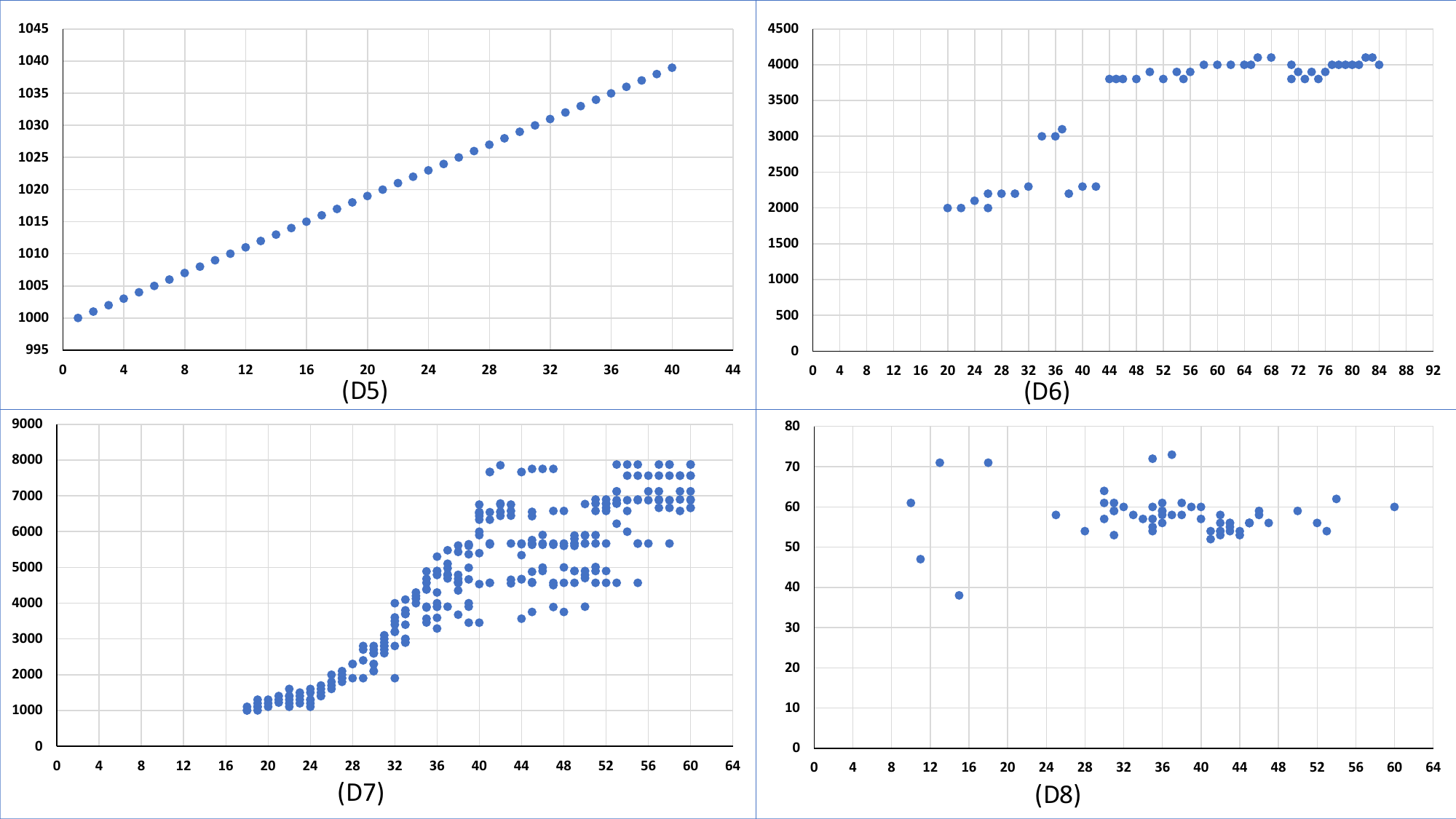}
\caption{Graphs for datasets D5 to D8.}
\label{fig2}
\end{figure}

\begin{figure}
\centering
\includegraphics[width=\textwidth]{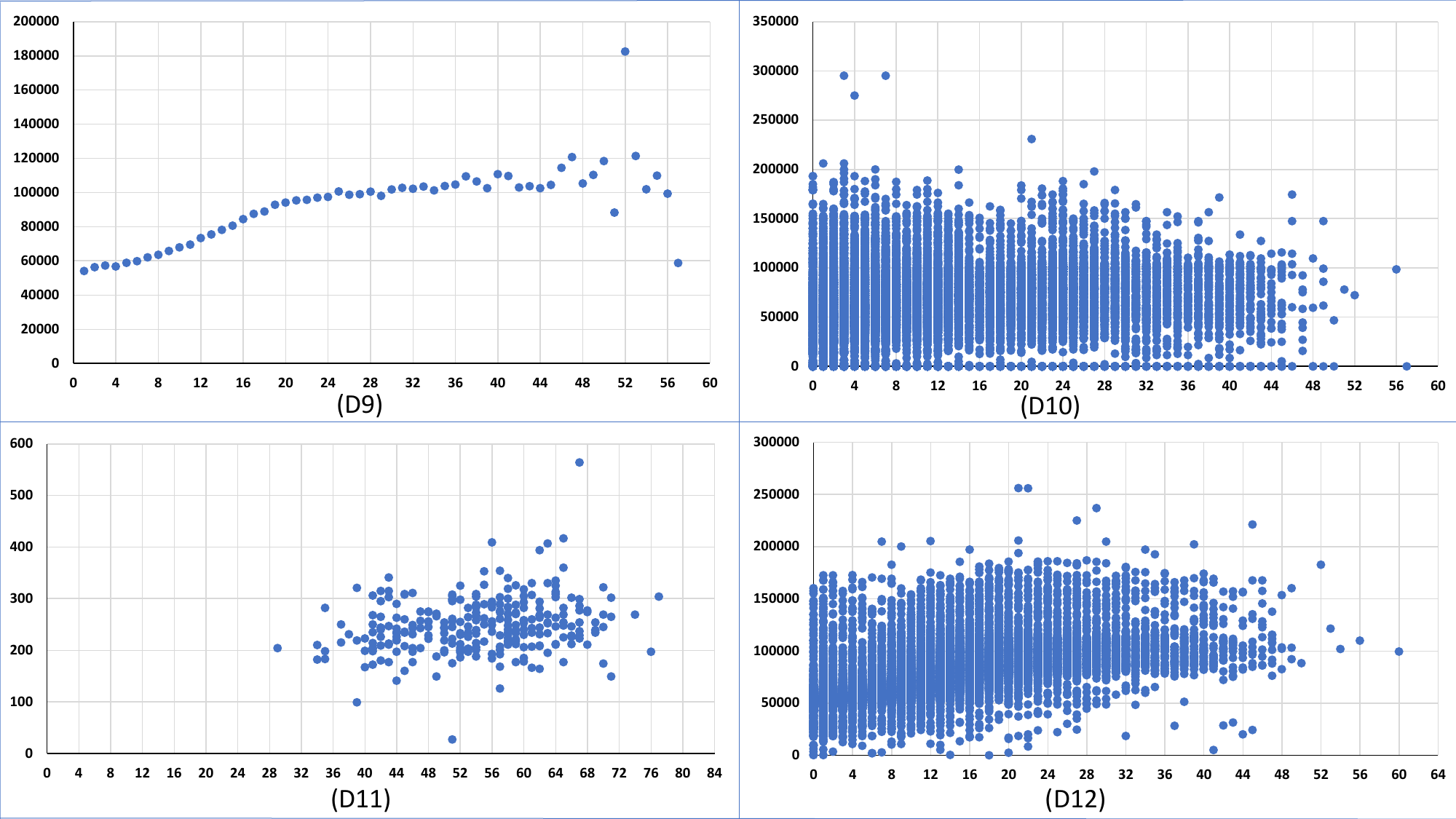}
\caption{Graphs for datasets D9 to D12.}
\label{fig3}
\end{figure}

To understand how humans partition numerical factors, we designed a series of graphs and asked several experts to partition the data given in the graphs. Initially, to produce a diverse collection of graphs with different data points, a set of graphs was shared and discussed with our own research team.  The team consists of three early-stage researchers and one senior researcher. These graphs include step functions, linear functions, and mixed data graphs. Finally, twelve graphs were selected to be shared with humans (see Figs.~\ref{fig1},~\ref{fig2} and~\ref{fig3}). These graphs are obtained from nine synthetic datasets (D1 to D9) and three real-world datasets (D10 to D12). These synthetic datasets (D1 to D9) have only two numerical attributes. 
The dataset D10 is a real-world dataset. The data set, DC public government employees~\cite{D_public}, contains 33,424 records of DC public government employees and their salaries in 2011. This dataset is sourced from the Washington Times via Freedom of Information Act (FOIA) requests.
The dataset D11 is Heart disease dataset~\cite{heartds}, and it is sourced from the UCI machine learning repository. This dataset has 13 attributes and 303 records. We used attribute~\emph\{Age\} and~\emph\{Cholesterol\} for drawing the graph. 
The dataset D12 is a New Jersey (NJ) school teacher
salaries (2016)~\cite{NJ_teacher} sourced from the (NJ) Department of Education. It contains 138715 records and 15 attributes. We have taken only an initial 23000 rows from the dataset. We are interested in the column~\emph\{experience\_total\} and~\emph\{salary\}. 
A copy of all these datasets is available in the GitHub repository~\cite{realdata}.

\begin{table}
\caption{The comparison of human perception to identify number of partitions based on their profile.}
\label{tb1}
\begin{center}
\begin{tabular}{|p{1.2cm}|p{3cm}|p{3.3cm}|p{3cm}|}
\hline
\multirow{2}{*}{Datasets} & \multirow{2}{*}{Number of Partitions} & Total responses from DS/ML experts $= 60\%$ &Total Responses from Non-expert People $=40\%$ \\\cline{3-4} 
 &  & \% Responses &   \% Responses\\\hline
\multirow{2}{*}{D1}  & 2  & 93.3\%  & 90\%\\
    & 3  &  6.67\% & 10\% \\\hline
\multirow{2}{*}{D2}  & 2  &  73.3\% & 60\% \\
    & 3  &  26.6\% & 40\%\\\hline
\multirow{3}{*}{D3} & 2 & 0\% & 0\% \\
    & 3 & 93.3\%  & 100\% \\
    & 4 & 6.66\% & 0\%\\\hline
\multirow{3}{*}{D4}  & 0 & 33.3\%  & 20\%\\
    & 2  & 53\% & 70\%\\
    & 3  & 13.3\%  & 10\%\\\hline
\multirow{3}{*}{D5}  & 0 & 93.3\% & 90\%\\
    & 2  &  0\% &0\%\\
    & 3  & 6.6\%  & 10\%\\\hline
\multirow{4}{*}{D6}  & 2  & 13.3\% & 30\%\\
    & 3  & 26.6\% & 0\%\\
    & 4  & 26.6\% & 40\%\\
    & 5  & 33.3\% & 30\%\\\hline
\multirow{3}{*}{D7}  & 2  & 60\%   & 40\%\\
    & 3  & 20\%   & 40\%\\
    & 4  & 20\%   & 20\%\\\hline
\multirow{3}{*}{D8}  & 2  & 33.3\% & 60\%\\
    & 3  & 66.6\% & 30\%\\
    & 4  & 0\%    & 10\%\\\hline
    
\multirow{3}{*}{D9}  &  2  &  73.3\% & 60\% \\
    &  3  & 26.6\%  & 30\% \\
    &  4  &  0\%    & 10\% \\\hline
    
\multirow{5}{*}{D10} & 0  & 40\%   & 40\%  \\
    &  2  & 40\%   & 30\% \\
    &  3  & 6.66\% & 20\% \\
    &  4  & 6.66\% & 0\% \\
    &  5  & 6.66\% & 10\% \\\hline

\multirow{4}{*}{D11} &  0  & 53.3\% & 60\%  \\
    &  2  & 26.6\% & 30\% \\
    &  3  & 20\%   & 0\% \\
    &  4  &  0\%   & 10\% \\ \hline

\multirow{5}{*}{D12} & 0  & 40\%   & 20\%  \\
    &  2  &  26.6\%& 30\% \\
    &  3  & 6.66\% & 30\% \\
    &  4  &  20\%  & 10\%  \\
    &  5  &  6.66\%& 10\%\\\hline

\hline
\end{tabular}
\end{center}
\end{table}

\begin{table}
\caption{The comparison of human perceived cut-points with the LSQM and LADM measures.}
\label{tb2}
\begin{center}
\begin{tabular}{|p{0.7cm}|p{0.3cm}|p{0.7cm}|p{6.2cm}|p{2.1cm}|p{2.1cm}|}
\hline
 \multirow{2}{*}{D} & \multirow{2}{*}{P} & \multicolumn{2}{c|}{Human Perception} & LSQM & LADM\\
 \cline{3-6} 
& & R & Approx. near Cut-Points & Cut-points  & Cut-Points \\\hline
D1 & 2 & 92\% & 50(91.3\%), 48(8.6\%)  & 50 & 50\\
   & 3 & 8\% & (48,60)(50\%), (20,50)(50\%)  & (20, 50) & (20, 50)\\\hline
D2 & 2 & 68\% & 50(88.2\%), 52(11.7\%)  & 52 & 52 \\
   & 3 & 32\% & (50,54)(37.5\%), (20,53)(25\%) & (52, 54) & (52, 54) \\\hline
D3 & 3 & 96\% & (32,52)(62\%), (30,52)(16.6\%) & (32,52) & (32,52) \\
   & 4 & 4\% & (20,32,52)(100\%)  & (32,52,55) & (32,52,60)\\\hline
D4 & 0 & 28\% & NA &  NA & NA \\     
   & 2 & 60\% & 20(86.6\%), 25(13.3\%)  &20 & 20 \\
   & 3 & 12\% & (20,45)(66.6\%), (20,30)(33.3\%)  & (12, 24) & (12, 25) \\\hline
D5 & 0 & 92\% & NA   & NA & NA\\      
   & 2 & 0\% & 0\%  & 20 & 19 \\
   & 3  & 8\% & (14,28)(100\%) & (13, 26) & (13, 26) \\\hline
D6 & 2 & 20\% & 32(40\%), 42(40\%) 50(20\%)  & 42 & 42 \\
   & 3 & 16\% & (42,68)(50\%), (32,42)(25\%)  & (32, 42) & (32, 42) \\
   & 4  & 32\% & (32,37,42)(87.5\%), (33,37,43)(12.5\%) & (32, 37, 42) & (32, 37, 42)\\
   & 5 & 32\% & (32,42,37,68)(87.5\%), (17,32,38,42)(12.5\%)  & (32, 37, 42, 56) &  (32, 37, 42, 56) \\\hline
     
D7 & 2  & 52\%  & 40(84.6\%), 50(7.6\%), 36(7.6\%)& 35 & 33 \\
   & 3  & 28\% & (32,39)(57.1\%) & (32, 39) & (32, 39) \\
   & 4 & 20\% &(32,39,50)(60\%), (41,47,53)(40\%)  & (32,39,52) & (32,39,52) \\\hline
    
 D8 & 2 & 44\% & 18(36\%), 30(27\%)  & 40 & 40 \\
    & 3 & 52\% & (28,47)(53.8\%), (18,47)(23\%) & (13, 15) & (40,45) \\
    & 4 & 4\% &(18,47,54)(100\%)  & (11, 13, 15) & (13,15,18)\\ \hline
 D9 & 2 & 68\% & 40(41\%), 50(23.5\%),47(23.5\%) & 15 & 13 \\
    & 3 &  28\% & (24,36)(57\%), (36,47)(28.5\%) &  (14, 50) & (8,15)\\ 
    & 4 & 4\% & (24,39,47)(100\%)  & (14,50,52) & (13,15,18) \\\hline
   
 D10  & 0 & 40\% & NA   &  NA & NA\\
      & 2 & 36\% & 44(33.3\%), 24(33.3\%), 52(22\%)  & 56 & 11  \\
      & 3 &  12\% & (20,32)(66.6\%), (18,45)(33.3\%)   & (11,56) & (11,56)\\
      & 4  & 4\%  & (12,29,42)(100\%)&  (49,50,56) & (11,52,56) \\
      & 5 & 8\% & (12,24,30,40)(100\%)  &  (11,49,50,56) & (11,41,50,56)\\\hline
D11   & 0 & 56\% & NA   &  NA & NA\\
      &  2 & 28\% & 52(42.8\%), 60(42.8\%), 67(14\%)   & 67 & 67 \\
      & 3 & 12\% & (48,68)(66.6\%), (40,68)(33.3\%)  & (67,70) & (67,70) \\
      & 4  & 4\% & (40,48,68)(100\%) & (51, 63, 67) & (62,67,70) \\\hline
D12 & 0 & 32\% & NA  &  NA & NA\\
    & 2 & 28\% & 50(42.8\%), 40(28.5\%), 24(28.5\%)  & 17 & 15\\
    & 3  & 16\% & (22,32)(50\%), (14,34)(25\%), (27,44)(25\%)  & (15,51) & (14,38) \\
    & 4 & 16\% & (9,31,58)(50\%), (20,36,48)(25\%), (10,20,30)(25\%)         & (15,51,52) & (10,17,38) \\
    & 5  & 8\% &  (16,28,36,44)(50\%), (7,20,28,36)(50\%)  &  (15,50,51,52) & (14,37,51,52) \\

\hline
\multicolumn{6}{c}{D: Datasets; P: number of partitions; R: percentage of responses}
\end{tabular}
\end{center}
\end{table}

We designed a Google form by providing a series of graphs containing different types of numerical data points and relevant questions to collect human responses and their perceptions about discretization. We put some constraints in the Google form to know whether a response is submitted by DS/ML experts or not.  By employing this procedure, we compare and comprehend the perceptions of both DS/ML experts and non-expert responders. 

The Google form was sent to fifty DS/ML experts and non-experts to estimate the number of partitions and the ranges of these partitions to obtain the cut-points. 
The following data was gathered and compiled from the experiments: respondent identification (name), their email addresses, domain expertise (DS/ML expert or non-expert), number of partitions identified, and ranges of each partition.



\section{Analysis and Result}
\label{sec6}
Out of the fifty responses received via the Google form, two were incomplete; therefore, we did not consider them for the analysis. From the rest of the forty-eight responses, we divided the responses into two categories: expert responses and non-expert responses.

\begin{table}[!t]
\caption{Similarity between human perceived, LSQM and LADM cut-points.}
\label{tb3}
\centering
\begin{tabular}{|p{0.6cm}|p{0.4cm}|p{2cm}|p{2cm}|p{3cm}|p{1.6cm}|p{1.4cm}|}
\hline
D & P & LSQM Cut- Points & LADM Cut- Points & Human Perceived Cut-Points (Near to LSQM Cut-Points) &Matching\% & Matching status \\\hline

D1 & 2 & 50 & 50 & 50 (91.3\%)  & 91.3\%  & Very High\\  
   & 3 & (20,50) & (20,50) & (20,50)(50\%) &  50\% &   Medium \\\hline
     
D2 & 2 & 52 & 52 &  52(11.7\%) &  11.7\% & Low\\
   & 3 & (52,54) & (52,54) & (50,54)(37.5\%)  & 19\% & Low\\\hline
   
D3 & 3 & (32,52) & (32,52) & (32,52)(62\%)  & 62\%  & High \\
   & 4 & (32,52,55) &(32,52,60) & (20,32,52)(100\%)  & 59\% & Medium  \\\hline
  
D4 & 2 & 20 & 20 & 20(80.6\%) &  80.6\% & Very High\\
   & 3 & (12,24) & (12,25) & (20,30)(33.3\%)  & 0\% & No match\\\hline
  
D5 & 3 & (13,26) & (13,26) & (13,26)(100\%) & 100\% &   Very High \\\hline
 
D6 & 2 & 42 & 42 & 42(40\%)  & 40\% & Medium\\
   & 3 & (32,42) & (32,42) & (32,42)(25\%)&  25\% & Low\\
   & 4 & (32,37,42) & (32,37,42) & (32,37,42)(85.7\%) &  85.7\% & Very High\\
   & 5 & (32,37,42,56) &(32,37,42,56) & (32,37,42,68)(85.7\%)  & 75\% &  High\\\hline
 
D7 & 2 & 35 & 33 & 36(7.6\%) & 0\% & No match\\
   & 3 & (32,39) & (32,39) & (32,39)(57\%)  & 57\% & High\\
   & 4 & (32,39,52) & (32,39,52) & (32,39,52)(60\%)  & 60\% & High\\\hline
  
D8 & 2 & 40 & 40 & 30(27\%) & 0\% &  No match\\
   & 3 & (13,15) & (40,45) & (18,47)(23\%)  & 0\%  & No match\\
   & 4 & (11,13,15) & (13,15,18) &(18,47,54)(100\%) &  0\% & No match\\\hline
   
D9 &  2 & 15  & 13 & 40(41\%)  & 0\%  &  No match \\
   &  3 & (14,50) & (8,15)& (24,36)(57\%)   & 0\%  & No match   \\
   &  4 & (14,50,52) & (10,17,33)  & (24,39,47)(100\%) &  0\% & No match  \\\hline
   
D10 & 2  & 56 & 11 & 52(22\%) & 0\%  &  No match \\
    & 3 & (11,56) & (11,56) & (18,45)(33.3\%) & 0\%  &  No match \\
    & 4 & (49,50,56) & (11,52,56)& (12,29,42)(100\%)  &  0\% & No match  \\
    & 5 & (11,49,50,56) & (11,41,50,56) & (12,24,30,40)(100\%) &  0\%  & No match  \\\hline
D11 & 2  & 67 & 67 & 50(42.8\%)   & 0\%  & No match  \\
    & 3 & (67,70) & (67,70)  & (48,68)(66.6\%)   & 0\%  & No match  \\
    & 4 & (51,63,67) & (62,67,70) & (40,48,68)(100\%)  & 0\%  &  No match \\\hline

D12 & 2 & 17 & 15 & 24(28.5\%) &  0\%  & No match  \\
    & 3 & (15,51) & (14,38) & (14,34)(25\%)   & 0\%  & No match  \\
    & 4 & (15,51,52) & (10,17,38)  & (20,36,48)(25\%)  & 0\%  & No match  \\
    & 5  & (15,50,51,52) & (14,37,51,52) & (16,28,36,44)(50\%) & 0\%  & No match\\\hline
\multicolumn{6}{c}{D: Datasets; P: number of partitions; R: percentage of responses}\\
\multicolumn{6}{c}{Very High: 80-100\%, High:60-80\%, Medium:40-60\%, Low:1-40\%,  No match: 0\%}\\
\end{tabular}
\end{table}

\begin{table}[!t]
\caption{Analysis of unmatched datasets in regard of number of partitions ($\#$) for LSQM, LADM and human perceived cut-points.}
\label{tb4}
\centering
\begin{tabular}{|p{0.6cm}|p{0.4cm}|p{1.8cm}|p{0.8cm} |p{1.8cm}|p{0.8cm}|p{1.9cm}|p{0.8cm}|p{3.3cm}|}
\hline
 \multirow{2}{*}{D} &  
 \multirow{2}{*}{ P}
 
 & \multicolumn{2}{l}{LSQM Method} & \multicolumn{2}{|l}{LADM Method} & \multicolumn{2}{|l|}{Human Perception} & \multirow{2}{*}{Remarks} \\\cline{3-8} 
 & & LSQM cut-points & LC & LADM cut-points & LC & Human Perceived cut-points & LC & \\\hline
D8 & 2 & 40 & Yes & 40 & Yes & 30 & Yes & Matter of perception.\\
   & 3 & (13,15) & No &(40,45) & No & (18,47) & Yes & LSQM, LADM to be improved.\\
   & 4 & (11,13,15) & No & (13,15,18) & No & (18,47,54) & Yes &  LSQM, LADM to be improved.\\\hline
D9 & 2 & 15 & No & 13 & No & 40 & Yes & LSQM, LADM need to be improved.\\
   & 3 & (14,50) & Yes & (8,15) & No & (24,36) & Yes & Matter of perception. However, LADM needs to be improved.\\
   & 4 & (14,50,52) & No & (10,17,33) & No & (24,39,47) & Yes & LSQM and LADM to be improved.\\\hline

D10 & 2 &  56 & No & 11 & No  & 52 & No    & This dataset is an exceptional case; random cutpoints are obtained.     
 \\
    & 3 & (11,56) & No  & (11,56) & No & (18,45) &  Yes &   \\
    & 4 & (49,59,56) & No & (11,52,56) & No & (12,29,42) & Yes &  \\
    & 5 & (11,49,50,56) & No & (11,41,50,56) & No & (12,24,30,40) & Yes & \\ \hline
D11 & 2 & 67 & Yes & 67 & Yes & 50 & Yes & Matter of perception.\\
    & 3 & (67,70) & No & (67,70) & No & (48,68) & Yes & LSQM to be improved.\\
    & 4 & (51,63,67) & Yes & (62,67,70) & No & (40,48,68) & Yes & Matter of perception. However, LADM needs to be improved.\\\hline
D12 & 2 & 17 & Yes & 15 & Yes & 24 & Yes  &  This dataset is an exceptional case; random cutpoints are obtained.  \\
    & 3 & (15,51) & Yes & (14,38) & Yes  & (14,34) & Yes    &    \\
    & 4 & (15,51,52) &  No  & (10,17,38) & No & (20,36,48) & Yes  &   \\
    & 5 & (15,50,51,52) & No  & (14,37,51,52) & No & (16,28,36,44) & Yes   &  \\\hline
D4 & 3 & (12,24) & Yes & (12,25) & Yes& (20,30) & Yes & Matter of perception.\\\hline
D7 & 2 & 35 & Yes & 33 &  Yes & 36 & Yes & Matter of perception.\\
\hline
\multicolumn{8}{c}{D: Datasets; P: number of partitions; LC: Logical Correctness}\\
\end{tabular}
\end{table}

Table~\ref{tb1} illustrates the comparison of human perception to identify the number of partitions between the DS/ML experts' responses and non-expert people. We received $60\%$ responses from DS/ML experts and $40\%$ of answers from non-expert people.
We analyzed that responses from both categories were opposite for graph D8. Out of the total responses for D8, $33.3\%$ responses of DS/ML experts marked two partitions and $66.6\%$ responses of experts marked three partitions; however, $60\%$ of non-experts marked two partitions, and only $30\%$ marked three partitions.

In graphs D3 and D5, we analyzed that no contributor (experts or non-experts) marked two partitions. No non-expert contributors marked three partitions for graph D6 and four partitions for graph D3, whereas $26.6\%$ of DS/ML experts identified three partitions for D6, and $6.66\%$ experts marked four partitions in graph D3.
For graph D10, $40\%$ DS/ML experts and $40\%$ non-experts have marked no partition.

Table~\ref{tb2} illustrates the comparison between the results of human perception, the~\emph{LSQM} and the~\emph{LADM} measure. In the initial four columns, we detailed the dataset used, the count of partitions, the response percentage, and the approximate cut-points as observed by contributors. The last two columns present the cut-points assessed by the measures. 
Table~\ref{tb3} describes the similarity percentage between cut-points provided by human perceived experiment outcome and the~\emph{LSQM} and the~\emph{LADM} measures outputs. We have mentioned the cut-points from responses near the~\emph{LSQM} and the~\emph{LADM} provided cut-points. We determine the matching status by distributing the matching percentage into the following categories: VH (Very High), H (High), M (Medium), L (Low) and NM (No match). The distribution of ranges is mentioned at the bottom of Table~\ref{tb3}. 
It is clear from Table~\ref{tb3} that human perceived cut-points and the cut-points identified by the proposed measures~\emph{LSQM} and ~\emph{LADM} do not match for the datasets D8 to D12. 
In Table~\ref{tb4}, we present an analysis and reason for not getting similar cut-points for the datasets D8 to D12.
If we look at Fig.~\ref{fig1}(D8), then it seems logical to have cut-points at the data points of $40$ (LSQM, LADM cut-point) and $30$ (Human perceived cut-point) for two partitions on the X-axis. Humans divided the scattered points into the first partition and dense data points into the second partition. In contrast, both measures calculated the cut-point in the middle of the dense data points. This case can be observed as a matter of perception for human perceived cut-points, while the cut-points marked by both the measures seem analytically correct.
For the cases of three partitions and four partitions, human perceived cut-points $(18,47)$ and $(18,47,54)$ are good, but the cut-points provided by the~\emph{LSQM}measure and the~\emph{LADM} measure are not satisfactory.
 
Similarly, in dataset D9, human perception identified a single cut-point at $40$ for two partitions and two cut-points at $(24,36)$ for three partitions, which intuitively makes sense. 
However,\emph{LSQM} and \emph{LADM} produced cut-points at $15$ and $13$ for two partitions, which are not analytically accurate. On the other hand,\emph{LSQM} cut-points $(14,50)$ align analytically, albeit it remains a matter of perception. 
 Even though the three partitions provided by\emph{LADM} $(8,15)$ and the four partitions by~\emph{LSQM} $(14,50, 52)$ and~\emph{LADM} $(10,17,33)$ might seem illogical, human-perceived cut-points $(24,39,47)$ appear appropriate.

In dataset D11, the situation is again contingent on perception, with discrepancies arising for both two partitions and four partitions in the case of~\emph{LSQM}. For four partitions,~\emph{LADM} suggests cut-points $(62,67,70)$, which lack logical consistency. Meanwhile, for three partitions, both~\emph{LSQM} and ~\emph{LADM} present unexpected cut-points $(67,70)$.

Datasets D4 and D7 also lack matching results for three partitions and two partitions, respectively. In the case of D4, both~\emph{LSQM} and~\emph{LADM} suggest cut-points $(12,24)$ and $(12,25)$, while human perception identifies $(20,30)$ as the appropriate cut-points. This instance can be attributed to varying perceptions.

Similarly, for D7, the proposed cut-points by~\emph{LSQM} and~\emph{LADM} for two partitions are $35$ and $33$, respectively, which do not exactly align with the human-perceived cut-point of $36$. However, given the scattered distribution of data points on the graph, the difference between the proposed measures' cut-points and the human-perceived cut-point is negligible. In this case, both sets of cut-points can be considered suitable, further emphasizing the role of perception. While these cut-points do not match precisely, it does not affect the correctness of the measures due to the lack of similarity.

For datasets D10 and D12, the responses from contributors present a unique challenge. In the case of D10, $40\%$ of contributors indicated no partition, while the remaining contributors marked random cut-points for two, three, four, and five partitions. Similarly, for D12, $32\%$ of contributors opted for no partition, while $68\%$ of contributors designated random cut-points for various partitions. These random cut-points identified by humans are not easily aligned with the cut-points derived from the proposed~\emph{LSQM} and~\emph{LADM} measures. Furthermore, these random cut-points lack analytical correctness.

As a result, for datasets with such characteristics, it becomes difficult for humans to identify the most appropriate partitions. The absence of clear patterns or logic in the random cut-points makes it challenging to establish meaningful partitions, emphasizing the complexity of the task in these scenarios.
 

\begin{figure}
\centering
\includegraphics[width=\textwidth]{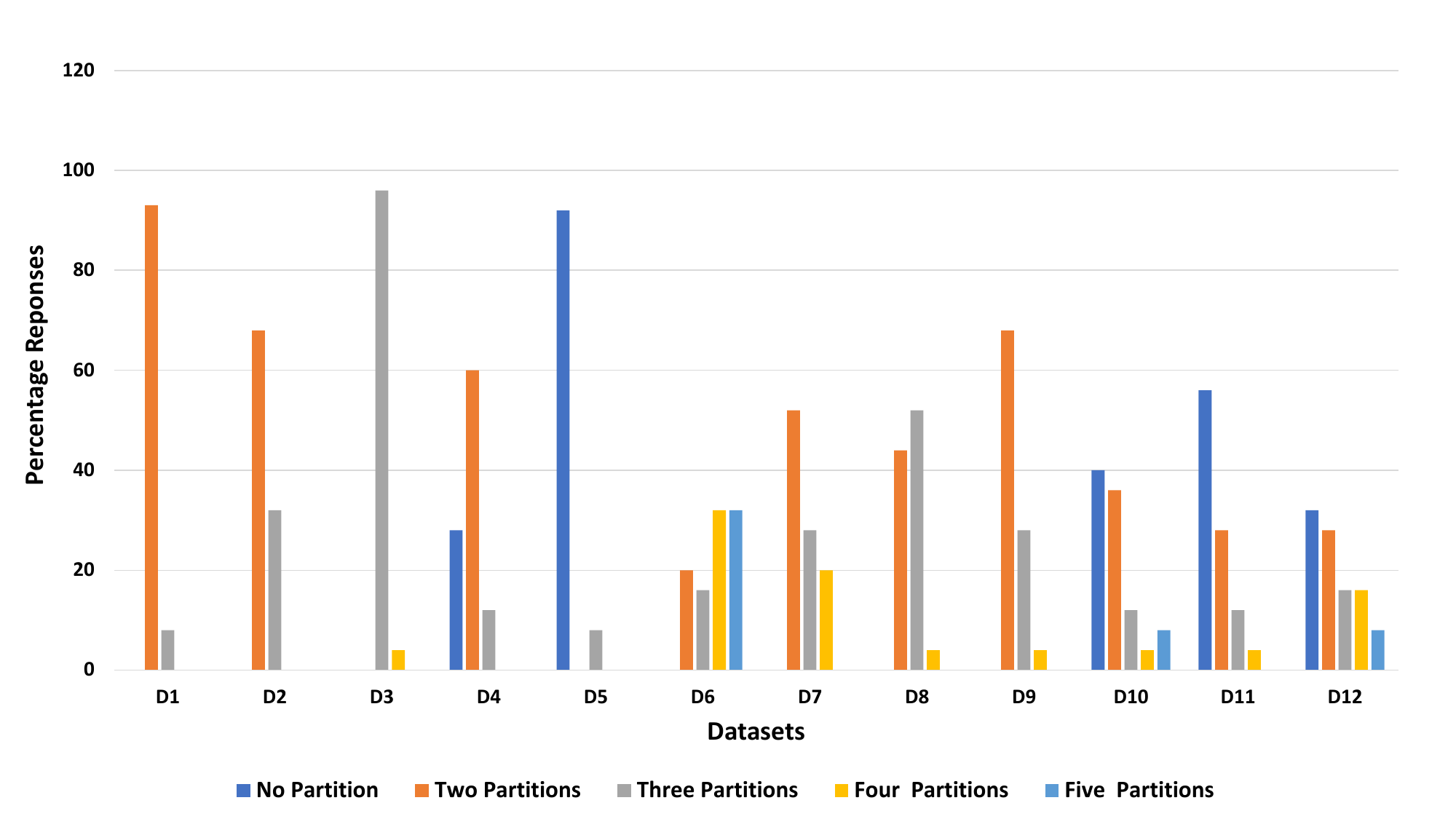}
\caption{Percentage Responses of partitions for each dataset.}
\label{fig4}
\end{figure}

The distribution of response percentages for each partition across the datasets is visually represented in Fig.~\ref{fig4}. Notably, datasets D5, D10, D11, and D12 exhibit a significant prevalence of responses indicating no partition compared to other partition numbers. Hence, it proves that for these specific datasets, human participants struggled to form a definitive perception.  It suggests that individuals had difficulty forming a clear and consistent perception of these datasets, making it unclear for them to identify appropriate cut-points.

It is worth highlighting that datasets D5 and D6 share a similar visual appearance. However, they received different responses in terms of cut-points. This discrepancy can be attributed to the distinct distribution of data points within each dataset. Notably, D6 did not receive any responses suggesting no partition, while D5 lacked responses suggesting two partitions.

Conversely, for datasets D1, D2, D4, D7, and D9, the majority of responses predominantly indicated the presence of two partitions. This indicates a higher degree of consensus among contributors regarding the presence of two partitions in these datasets.


Table~\ref{tb3} provides an overview of the alignment between human-perceived cut-points and those observed by the~\emph{LSQM} and~\emph{LADM} measures. The analysis reveals that $25\%$ of responses exhibited a ~\emph{Very High} level of similarity, $25\%$ demonstrated a ~\emph{High} level of similarity, $18.7\%$ displayed a ~\emph{Medium} level of similarity, and an additional $18.7\%$ showed a ~\emph{Low} level of similarity.
When considering the collective matching statuses, it becomes evident that approximately $68.7\%$ of the responses closely resembled the cut-points identified by the proposed~\emph{LSQM} and \emph{LADM} measures. This analysis primarily pertains to the initial datasets (D1 to D7), as random cut-points were observed in the responses for datasets D8 to D12. These random cut-points in the latter datasets presented challenges in aligning them with the analytically calculated cut-points generated by the proposed measures. Further details and explanations for the dissimilarity in cut-points for datasets D8 to D12 can be found in Table~\ref{tb4}.

\section{Conclusion}
\label{sec7}
This paper is the first step toward understanding the human perception regarding partitioning numerical attributes. We meticulously examine the partitions perceived by humans and compare them with the outputs generated by both proposed measures. 
Our approach involved evaluating human perception by presenting a series of graphs containing numerical data and subsequently comparing human-perceived cut-points for partitioning with the results generated by the~\emph{LSQM} and~\emph{LADM} measures.

The outcomes of this study indicate a close alignment between the cut-points produced by the proposed measures and those perceived by humans. Particularly for the initial datasets (D1 to D7), our proposed measures yielded results that closely approximated human perception. However, certain exceptional cases, such as datasets D10 and D12, highlighted situations where humans faced challenges identifying optimal partitions. The results also demonstrate that both measures yield approximately similar outcomes.
These findings represent a promising step forward, signifying progress in the pursuit of advancing ARM by identifying numerical attribute partitions that reflect best the impact of an independent numerical attribute on a dependent numerical attribute. In future research endeavors, we intend to explore inter-measures for comparing partitions with varying numbers of $k$-partitions.

\section*{Acknowledgements}
\label{sec8}
This work has been conducted in the project ``ICT programme" which was supported by the European Union through the European Social Fund.

\bibliographystyle{splncs04}
\bibliography{rr}

\end{document}